\documentclass[10pt,twocolumn,letterpaper]{article}

\usepackage{cvpr}              

\usepackage{graphicx}
\usepackage{amsmath}
\usepackage{amssymb}
\usepackage{booktabs}

\usepackage[pagebackref,breaklinks,colorlinks]{hyperref}

\usepackage{algorithm}
\usepackage{algorithmic}
\usepackage{bm}
\usepackage{amsmath}
\usepackage{bbm}
\usepackage{color}
\usepackage{amssymb}
\usepackage{subfloat}
\usepackage{multirow}

\usepackage[capitalize]{cleveref}
\crefname{section}{Sec.}{Secs.}
\Crefname{section}{Section}{Sections}
\Crefname{table}{Table}{Tables}
\crefname{table}{Tab.}{Tabs.}

\def\cvprPaperID{7413} 
\def\confName{CVPR}
\def\confYear{2022}

\begin{document}

\title{Entropy-based Active Learning for Object Detection \\with Progressive Diversity Constraint}

\author{Jiaxi Wu$^{1,2}$, Jiaxin Chen$^{2}$, Di Huang$^{1,2}$\footnotemark[1]\\
$^{1}$State Key Laboratory of Software Development Environment, Beihang University, Beijing, China\\
$^{2}$School of Computer Science and Engineering, Beihang University, Beijing, China\\
{\tt\small \{wujiaxi,jiaxinchen,dhuang\}@buaa.edu.cn}
}
\maketitle

\begin{abstract}
   Active learning is a promising alternative to alleviate the issue of high annotation cost in the computer vision tasks by consciously selecting more informative samples to label. Active learning for object detection is more challenging and existing efforts on it are relatively rare. In this paper, we propose a novel hybrid approach to address this problem, where the instance-level uncertainty and diversity are jointly considered in a bottom-up manner. To balance the computational complexity, the proposed approach is designed as a two-stage procedure. At the first stage, an Entropy-based Non-Maximum Suppression (ENMS) is presented to estimate the uncertainty of every image, which performs NMS according to the entropy in the feature space to remove predictions with redundant information gains. At the second stage, a diverse prototype (DivProto) strategy is explored to ensure the diversity across images by progressively converting it into the intra-class and inter-class diversities of the entropy-based class-specific prototypes. Extensive experiments are conducted on MS COCO and Pascal VOC, and the proposed approach achieves state of the art results and significantly outperforms the other counterparts, highlighting its superiority.
\end{abstract}

\section{Introduction}
\label{sec:intro}

\renewcommand{\thefootnote}{\fnsymbol{footnote}}
\footnotetext[1]{Corresponding author.}

\begin{figure*}[!t]
	\centering
	\includegraphics[width=0.9\linewidth]{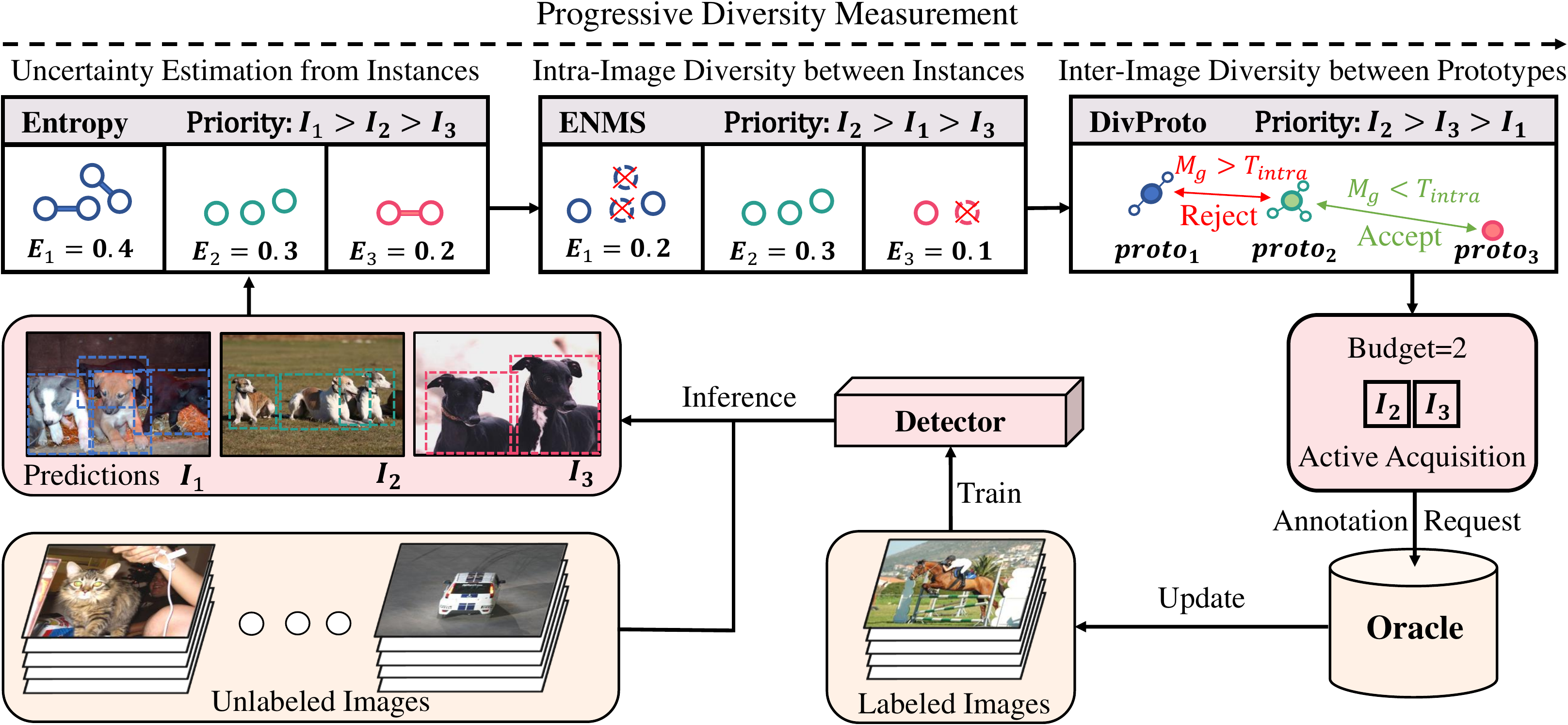}
	\caption{Framework overview. The hollow circles refer to uncertainty predictions and the solid ones denote the aggregated prototypes. At each cycle, the detector is trained with labeled images and infers the unlabeled ones. Instance uncertainty is first computed based on the entropy. ENMS then performs on each image to remove redundant instances. DivProto aggregates the instances of each image to prototypes and rejects the images close to the selected ones. The priority of active acquisition is illustrated with 3 examples: $I_1$, $I_2$, $I_3$. At the end of each cycle, the selected images (\emph{e.g.} $I_2$, $I_3$) are labeled by an oracle.}
	\label{fig:method}
\end{figure*}

During the past decade, visual object detection~\cite{fasterr-cnn,fcos} has been greatly advanced by deep Convolutional Neural Networks (CNN)~\cite{vgg,resnet} with persistently increasing performance reported. Unfortunately, strong CNNs generally make use of huge amounts of annotated data to fit extensive numbers of parameters, and training such detectors requires bounding-box labels on images, which is quite expensive and time-consuming. As one of the most promising alternatives to alleviate this dilemma, active learning~\cite{coreset,sraal} aims to reduce this high cost by consciously selecting more informative samples to label, and it is expected to deliver a higher accuracy with much fewer annotated images compared to that conducted in the random way.

In the community of computer vision, active learning is mainly discussed on image classification~\cite{multiclass,coreset,vaal}, where current methods roughly go into two categories, \emph{i.e.} uncertainty-based ~\cite{dbal,ll4al} and diversity-based~\cite{preclustering,coreset}. Uncertainty-based methods~\cite{dbal,ll4al} screen informative samples from entire databases according to their ambiguities~\cite{multiclass,ll4al,ensemble, dbal}. As the samples are separately predicted, they are efficient but tend to incur high correlations. Diversity-based methods~\cite{preclustering,coreset,cdal} claim that informative samples are the representatives of the whole data distribution and identify a subset using distance metric~\cite{coreset} or class probability~\cite{cdal}. They prove effective for small models, but suffer from high computational complexity. In addition, there exists another trend to combine the uncertainty- and diversity-based methods as hybrid ones ~\cite{convex,effectiveloss,divgradient}, and the achieved superiority figures out a promising alternative to other tasks.

As we know, object detection is more complicated than image classification, where object category and location are simultaneously output. In this case, active learning is desired to deal with various numbers of objects within images and the essential issue is to make image-level decisions according to instance-level predictions. The diversity-based method, CDAL~\cite{cdal}, applies spatial pooling to roughly approximate instance aggregation and formulates image selection as a reinforcement learning process. Regarding uncertainty-based methods, Learn Loss~\cite{ll4al} designs a task-free loss prediction module, and computes the image uncertainty by image-level features instead of instance-level ones, while MIAL~\cite{mial} defines the image uncertainty as that of the top-$K$ instances and estimates it with multiple instance learning based re-weighting. Since the diversity-based methods do not fully make use of categorical information and the uncertainty-based ones do not well measure the discrepancy of informative samples, the two types of methods leave room for performance improvement.

In this study, we propose a novel hybrid approach to active learning for object detection, which considers both the uncertainty and diversity at the instance level. To balance the computational complexity, the proposed approach works in a two-stage manner, as Fig.~\ref{fig:method} displays. At the first stage, we estimate the uncertainty of each image by an Entropy-based Non-Maximum Suppression (ENMS). ENMS performs Non-Maximum Suppression on the calculated entropy in the feature space to remove instances that bring redundant information gains, where a bigger value of the entropy refined by ENMS indicates the selection priority of an unlabeled image. At the second stage, unlike existing uncertainty-based methods~\cite{bmvc18,mial} which choose the top-$K$ images for annotation, we introduce the diverse prototype (DivProto) strategy to ensure instance-level diversity across images. It employs the prototypes~\cite{protofs,daodgpa} as the image-level representatives by aggregating the class-specific instances, and decomposes the cross-image diversity into the intra-class and inter-class ones. We then acquire the images of the minority classes for inter-class diversity and reject the ones that incur redundancy for intra-class diversity. In this way, the proposed approach combines the advantages of the uncertainty and diversity based ones in a bottom-up manner. We evaluate the proposed approach on MS COCO~\cite{coco} and Pascal VOC~\cite{voc07,voc12} and deliver state of the art scores on both of them, highlighting its effectiveness.

\section{Related Work}

\subsection{Active Learning on Image Classification}
As stated, the majority of the studies on active learning in computer vision target on image classification and are mainly categorized into diversity-based~\cite{coreset, cdal} and uncertainty-based~\cite{dbal, ll4al} ones.

The diversity-based methods screen a subset of samples to represent the global distribution by clustering~\cite{preclustering} or matrix partition~\cite{matricpartition} techniques. Core-set~\cite{coreset} defines active learning as a core-set selection problem and adopts $k$-center approximation. To improve efficiency, CDAL~\cite{cdal} replaces distance based similarity with the KL divergence. Those methods are theoretically complete but computationally inefficient when dealing with high-dimensional data. 

The uncertainty-based methods select ambiguous samples which are regarded as the most informative to the entire dataset~\cite{multiclass, dbal, ensemble, ll4al}. Many efforts are made to estimate the data uncertainty, \emph{e.g.} the entropy of class posterior probabilities~\cite{multiclass}. In this case, \cite{dbal} introduces Bayesian CNNs as an expert; \cite{ensemble} employs deep ensembles and Monte-Carlo dropout; and Learn Loss~\cite{ll4al} proposes a task-free image-level loss prediction module. The methods above are efficient, but bring in redundant samples for annotation.


Some alternatives~\cite{hybrid_trans, convex, seguncertainrepresent} combine the advantages of both types. With the uncertainty and diversity scores, \cite{convex} simply chooses the minimal; \cite{effectiveloss} emphasizes the diversity at early cycles and moves to the uncertainty gradually; \cite{albl} views the fusion as a multi-armed bandit problem and re-weights different scores. VAAL~\cite{vaal} performs uncertainty estimation on whether a data point belongs to the labeled or unlabeled pool, and acquires the samples most similar to the latter. SRAAL~\cite{sraal} further exploits the uncertainty estimator and the supervised learner to enclose annotation information. BADGE~\cite{divgradient} models the uncertainty and diversity by the gradient magnitudes and directions from the last layer respectively. The hybrid methods achieve promising results and suggest a new fashion for other tasks.


\subsection{Active Learning on Object Detection}
Object detection has been greatly progressed by CNNs~\cite{vgg, resnet} in the past few years mainly under the one-stage~\cite{ssd, retinanet, fcos} and two-stage~\cite{fastrcnn, fasterr-cnn} frameworks. As detection annotation is more expensive and time-consuming, active learning comes into focus in this branch, and preliminary attempts demonstrate its necessity \cite{visapp, hitl, alodmmd, bmvc18, locaware}. Meanwhile, with both object category and location to predict, this task is more challenging. 

Both the diversity-based and uncertainty-based methods have been recently adapted to object detection, and they extend direct image-level decision by integrating instance-level predictions. For the former, CDAL~\cite{cdal} represents the image using the detection features after spatial pooling to approximate this process. In spite of certain potential, a substantial performance gain requires global instance-level feature comparison, which incurs a vast complexity. For the latter, Learn Loss~\cite{ll4al} employs holistic image-level features for uncertainty estimation and with the task-free loss prediction module, it directly evaluates how much information an unlabeled image contributes. MIAL~\cite{mial} selects the image by measuring its uncertainty based on that of the top-$K$ instances re-weighted in a multiple instance learning framework, with noisy ones suppressed and representative ones highlighted. They ignore instance-level correlation within the whole data pool and thus deliver much redundancy. To address the issues above, this paper presents a way to jointly use their strengths to advance object detection.   

\section{Problem Statement}
This section starts with the formulation of active learning for object detection. The generic pipeline can be roughly grouped into three steps: (1) inference on unlabeled images with the existing detector, (2) image acquisition and annotation under a budget, and (3) detector training and evaluation on newly labeled images. These three steps execute in a loop and each iteration is viewed as a cycle (or stage). After each active learning cycle, the performance of the detectors represents the ability of the active acquisition methods since they select different images to annotate, where a fixed image amount is adopted as the annotation budget~\cite{ll4al, cdal, mial}. As detector training and evaluation are set in the same way, we focus on exploring a more effective acquisition method.



Suppose we have a large collection of candidate images $\{I_{i}\}_{i\in[n]}$ as well as a selected image set $\mathcal{S}=\{I_{s(j)}|s(j)\in[n]\}_{j\in[m]}$, where $[n]=\{1,\cdots,n\}$ and $[m]=\{1,\cdots,m\}$. Note that $\mathcal{S}$ denotes the labeled subset before each active acquisition cycle, which is initially chosen at random. 
Given a budget $b$, the batch active learning algorithm aims to acquire an image subset $\Delta{\mathcal{S}}$ in each cycle such that $|\Delta{\mathcal{S}}|=b$. $\Delta{\mathcal{S}}$ is subsequently labeled by an oracle, and is applied to update $\mathcal{S}$ as $\mathcal{S}:=\mathcal{S}\cup \Delta{\mathcal{S}}$. The oracle is requested to provide labels $\mathcal{Y}=\{y_{s(j)}\}_{j\in[m]}$ for each selected image. The learning model $D_{\mathcal{S}}$ is successively trained by $\mathcal{S}$ and $\mathcal{Y}$.

As depicted in Core-set~\cite{coreset}, the active learning problem is defined as minimizing the core-set loss $\sum\nolimits_{i\in[n]}l(I_{i},y_{i};D_{\mathcal{S}})$, where $y_{i}$ is the label of $I_{i}$. In the setting of object detection, the detector $D_{\mathcal{S}}$ is decomposed to an encoder $P_{\mathcal{S}}$ and a successive predictor $A_{\mathcal{S}}$. $P_{\mathcal{S}}$ encodes a set of spatial positions $\{pos_{k}\}_{k\in [t]}$ in $I_{i}$ to a set of features $P_{\mathcal{S}}(I_{i})=\{P_{\mathcal{S}}(I_{i},k)\}_{k\in[t]}$ by adopting the receptive fields~\cite{fasterr-cnn,ssd} or positional embeddings  \cite{detr} from $D_{\mathcal{S}}$.
Afterwards, $A_{\mathcal{S}}$ predicts $A_{\mathcal{S}}(P_{\mathcal{S}}(I_{i}))=\{\tilde{y}_{i,k}, c_{i,k},p_{i,k}\}_{k\in[t]}$ based on $P_{\mathcal{S}}(I_{i})$, where $\tilde{y}_{i,k}$, $c_{i,k}$ and $p_{i,k}$ are the predicted bounding box, object class and confidence score, respectively. The image-level core-set loss $l(\cdot)$ for object detection can be reformulated as: $\sum_{k\in[t]}l_{D}(P_{\mathcal{S}}(I_{i},k), {y}_{i,k}; A_{\mathcal{S}})$, where $l_{D}$ is the instance-level loss function.
To adopt the Core-set based solution, $l(\cdot)$ should be Lipschitz continuous as required by \emph{Theorem 1} in~\cite{coreset}. However, $P_{\mathcal{S}}(I_{i})$ is unordered, and thus is difficult to be explicitly defined, making $l(\cdot)$ not Lipschitz continuous. 



To address this issue, inspired by the uncertainty-based studies~\cite{bmvc18,mial}, we alternatively explore the empirical uncertainty from $P_{\mathcal{S}}(I_{i})$ and adopt the entropy-based formulation.
Specifically, we calculate the following entropy \cite{shannoentropy} for the $k$-th instance:
\begin{equation}
\mathbb{H}(I_{i},k)
=-p_{i,k}\log{p_{i,k}}-(1-p_{i,k})\log{(1-p_{i,k})},
\label{eq2}
\end{equation}
where $p_{i,k}$ is the confidence score predicted as the foreground of a certain category and $1-p_{i,k}$ as the background.

From Eq. \eqref{eq2}, the image-level \emph{basic detection entropy} is defined by replacing $l_{D}(\cdot)$ with $\mathbb{H}(I_{i},k)$ in $l(\cdot)$ as below:
\begin{equation}
\mathbb{H}(I_{i}|D_{\mathcal{S}})=\sum\nolimits_{k\in[t]}\mathbb{H}(I_{i}, k).\label{eq3}
\end{equation}

Based on $\mathbb{H}(I_{i}|D_{\mathcal{S}})$, the unlabeled images are sorted, and the top-$K$ images are selected as the acquisition set $\Delta {\mathcal{S}}$.

As visually similar bounding boxes contain redundant information which are not preferred when training robust detectors, it is desirable to select the most informative ones and abandon the rest. Moreover, such information redundancy occurs not only within each image but also across images, making it more difficult to retain the instance-level diversity. There still lacks a hybrid approach that considers the instance-level evaluation and achieves the image-level acquisition in the mean time.

\section{Method}



\subsection{A Hybrid Framework}\label{sec:framework}


In this subsection, we describe the details about our proposed hybrid framework, specifically designed for active learning on object detection.

As shown in Fig.~\ref{fig:method}, the proposed framework mainly consists of three modules: uncertainty estimation using the basic detection entropy, Entropy-based Non-Maximum Suppression (ENMS) and the diverse prototype (DivProto) strategy. The basic detection entropy in Eq.~\eqref{eq3} is adopted to quantitively measure the image-level uncertainty from object instances. ENMS is subsequently presented to remove redundant information according to the entropy, thus strengthening the instance-level diversity within images. DivProto further ensures the instance-level diversity across images by converting it into the inter-class and intra-class diversities formulated with class-specific prototypes. 

\begin{figure}[!t]
	\centering
	\includegraphics[height=4.7cm]{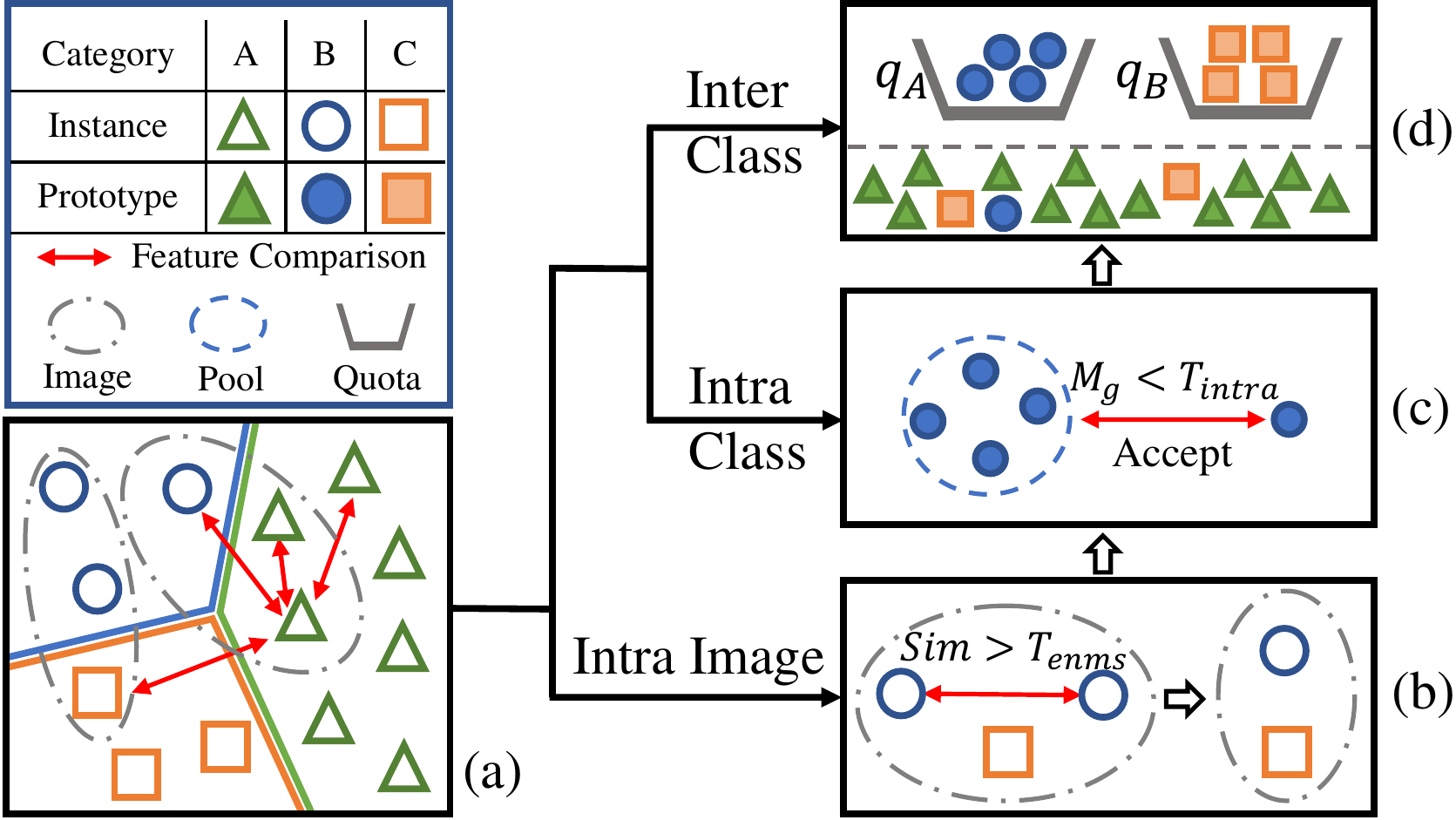}
	\caption{The hierarchy of the instance-level diversity is displayed in (a). (b) refers to the intra-image diversity by removing the instance-level redundancy via ENMS. (c) and (d) strengthen the intra-class and inter-class diversities across images formulated with the class-specific prototypes, respectively.}
	\label{fig:divmethod}
\end{figure}

To be specific, the hierarchy of our method for diversity enhancement is illustrated in Fig.~\ref{fig:divmethod}. 
The overall instance-level diversity (a) is divided into the intra-image diversity (b) via ENMS and the inter-image diversity is accomplished by DivProto, which is then decomposed into the inter-class and intra-class ones as shown in (c) and (d), respectively. By virtue of this progressive way, the diversity constraints on the predicted instances are effectively undertaken. The details are elaborated in the rest part.

\subsection{ENMS for Intra-Image Diversity}\label{sec:local}

\begin{algorithm}[!tb]
	\caption{Entropy-based Non-Maximum Suppression}
	\label{code:enms}
	\textbf{Input}: the predicted classes $\{c_{i,k}\}_{k\in[t]}$ \\
	                $~~~~~~~~~~~~~${the confidence scores $\{p_{i,k}\}_{k\in[t]}$} \\ 
	                $~~~~~~~~~~~~~$the instance-level features $\{\bm{f}_{i,k}\}_{k\in[t]}$ \\
	                $~~~~~~~~~~~~~$the threshold $T_{enms}$ (0.5 by default)\\
	\textbf{Output}: the image-level entropy $E_{i}$ after ENMS\\
	\textbf{Initialize}: $E_{i}:=0$
	\begin{algorithmic}[1] 
		\STATE Calculate the instance entropy $\{\mathbb{H}(I_{i},k)\}_{k\in [t]}$ according to Eq.~\eqref{eq2} 
		\STATE Initialize the indicating set $S_{ins}:=[t]$
		\WHILE{$S_{ins} \neq \varnothing$}
		\STATE Select the most informative instance $k_{pick}$ according to $k_{pick}:=\arg{\max_{k\in[S_{ins}]}{\mathbb{H}(I_{i},k)}}$ from $S_{ins}$ and update $S_{ins}:=S_{ins}-\{k_{pick}\}$
		\STATE Update the entropy $E_{i}:=E_{i}+\mathbb{H}(I_{i},k_{pick})$
		\FOR{$j$ in $S_{ins}$}
		\IF {$c_{i,j}=c_{i,k_{pick}}$ and ${\rm{Sim}}(\bm{f}_{i,j}, \bm{f}_{i,k_{pick}})>T_{enms}$}
		\STATE Remove the instance $j$ as $S_{ins}:=S_{ins}-\{j\}$
		\ENDIF
		\ENDFOR
		\ENDWHILE
	\end{algorithmic}
\end{algorithm}

As Eq.~\eqref{eq3} depicts, the basic detection entropy $\mathbb{H}(I_{i}|D_{\mathcal{S}})$ is a simple sum of the entropy of the candidate bounding boxes. Nevertheless, existing object detectors often generate a large amount of proposal bounding boxes with heavy overlaps, incurring severe spatial redundancy and high computational cost. This issue can be partially mitigated by applying Non-Maximum Suppression (NMS)~\cite{fasterr-cnn,ssd}, based on which bounding boxes belonging to the same instance are merged to a unified one. But NMS cannot deal with the instance-level redundancy, \emph{i.e.} instances with similar appearances presenting in the same context, which is supposed to be reduced in active acquisition. 


To overcome this shortcoming of NMS, we propose a simple yet effective \emph{Entropy-based Non-Maximum Suppression} (ENMS) as a successive step of NMS for instance-level redundancy removal. Specifically, we first compute the following Cosine distance $\rm{Sim}(\cdot,\cdot)$ to measure pair-wise inter-instance duplication: $\textrm{Sim}(\bm{f}_{i,k},\bm{f}_{i,j})=\frac{\bm{f}_{i,k}^{T}\cdot{\bm{f}_{i,j}}}{\left\|{\bm{f}_{i,k}}\right\|\left\|\bm{f}_{i,j}\right\|}$,
where $\bm{f}_{i,k}$ is the feature of the instance $k$ in the image $I_{i}$ extracted by $P_{\mathcal{S}}(\cdot)$. 
Subsequently, ENMS is performed on the indicating set $S_{ins}$ initialized as $[t]$, where $[t]$ is the set of all instances in $I_{i}$. As summarized in Algorithm~\ref{code:enms}, the basic idea of ENMS is to select the most informative instance $k_{pick}$ from $S_{ins}$ with the corresponding entropy $\mathbb{H}(I_{i},k_{pick})$ being accumulated for the image-level entropy $E_{i}$. Meanwhile, the remaining within-class instances that are similar to $k_{pick}$ (\emph{i.e.} the pair-wise similarity is larger than a threshold $T_{enms}$) are deemed as redundant ones, and further removed from $S_{ins}$. The procedure aforementioned is iteratively conducted until $S_{ins}$ becomes empty.


It is worth noting that ENMS only compares instances from the same categories w.r.t. the selected informative instance, and is thus computationally efficient. Meanwhile, ENMS extracts the instance-level features on-the-fly, which significantly reduces the memory cost. Additionally, ENMS can mitigate the unbalanced amount of instances per image, by means of redundant instance removal.

\subsection{Diverse Prototype for Inter-Image Diversity}\label{sec:global}

\begin{algorithm}[!tb]
	\caption{Diverse Prototype}
	\label{code:whole}
	\textbf{Input}: the labeled images $\mathcal{S}$\\ 
	$~~~~~~~~~~~~~$the unlabeled images $\{I_{i}\}_{i\in[n]}-\mathcal{S}$\\ 
	$~~~~~~~~~~~~~$the budget $b$ and the thresholds $T_{intra}$ and $T_{inter}$\\
	\textbf{Output}: the selected image set $\Delta{\mathcal{S}}$ to be labeled\\
	\textbf{Initialize}: $\Delta{\mathcal{S}}:=\varnothing$
	\begin{algorithmic}[1] 
		\STATE Calculate the entropy $\{E_{i}\}$ as well as the prototypes $\{\{\bm{proto}_{i,c}\}_{c\in[C]}\}$ for the set of the unlabeled images $\{I_{i}\}_{i\in[n]}-\mathcal{S}$ by ENMS and Eq.~\eqref{eq5}, respectively.
		\STATE Calculate the quotas $\{q_{c}\}_{c\in[C_{minor}]}$ based on $\mathcal{S}$
		\STATE Sort $\{I_{i}\}_{i\in[n]}-\mathcal{S}$ in descending order according to $\{E_{i}\}$
		\FOR{$i$ in $\left[\left|\{I_{i}\}_{i\in[n]}- \mathcal{S}\right|\right]$}
		\IF{${M}_{g}(I_{i},[C])<T_{intra}$ and ${M}_{p}(I_{i},[C_{minor}])>T_{inter}$}
		\STATE Select $I_{i}$ and update $\Delta{\mathcal{S}}:=\Delta{\mathcal{S}}\cup\{I_{i}\}$
		\FOR{$c$ in $[C_{minor}]$}
		\STATE Update $q_{c}:=q_{c}-1$ if $p(i,c)>T_{inter}$
		\STATE Update $C_{minor}:=C_{minor}-1$ if $q_{c}={0}$
		\ENDFOR
		\ENDIF
		\ENDFOR
		\STATE Fill up $\Delta{\mathcal{S}}$ with the rest images from the sorted set $\{I_{i}\}_{i\in[n]}- \mathcal{S}$ until $|\Delta{\mathcal{S}}|=b$
	\end{algorithmic}
\end{algorithm}

ENMS enhances the intra-image diversity, and the inter-image diversity, \emph{i.e.} the redundancy across images, still remains. Most conventional approaches~\cite{cdal} address this issue based on holistic image-level features, which are too coarse to fulfill instance-level processing in object detection. Some re-weighting based methods~\cite{albl,effectiveloss} can be adapted from image-level to instance-level, mitigating the inter-image redundancy. However, they need to compute the distances between all instance pairs, which incurs high memory and computational cost. Besides, current studies rarely consider the imbalanced classes of instances, making it hard to estimate the normalized diversity.



Inspired by the previous attempts~\cite{protofs,daodctf,daodgpa}, we introduce the prototypes to address the drawbacks above. Concretely, the $i$-th prototype of class $c$ is formulated as:
\begin{equation}
\bm{proto}_{i,c}=\frac{\sum\nolimits_{k\in[t]}{\mathbbm{1}(c,c_{i,k})\cdot{\mathbb{H}(I_{i},k)}\cdot{\bm{f}_{i,k}}}}{\sum\nolimits_{k\in[t]}{\mathbbm{1}(c,c_{i,k})\cdot{\mathbb{H}(I_{i},k)}}}, \label{eq5}
\end{equation}
where $\mathbbm{1}(c,c_{i,k})$ equals to $1$ if $c=c_{i,k}$, and $0$ otherwise. 


As shown in Eq. \eqref{eq5}, the prototype is formulated based on the entropy and the predicted class instead of the confidence score as in existing work \cite{daodgpa}, since our framework focuses on the information gain. Therefore, the instances with high classification confidence scores contribute less to the prototype than the uncertain ones.

Based on ENMS and the prototypes, we propose the \emph{Diverse Prototype} (DivProto) strategy to enhance the instance-level diversity across images. Specifically, we firstly sort the unlabeled images according to their entropy $\{E_{i}\}$ via ENMS in descending order. Subsequently, we improve the intra-class diversity via intra-class redundancy rejection as shown in Fig.~\ref{fig:divmethod}~(c) and the inter-class diversity via inter-class balancing as in Fig.~\ref{fig:divmethod}~(d). 

\textbf{Intra-class Diversity.} Given a candidate image $I_{i}$ and the prototypes of the acquired set $\Delta{\mathcal{S}}$, the intra-class diversity of $I_{i}$ is measured by the following metric:
\begin{equation}
\label{eq7}
M_{g}(I_{i}, [C])=\min\limits_{c\in[C]}{\max\limits_{j\in|{\Delta{\mathcal{S}}}|}{{\rm Sim}(\bm{proto}_{j,c}, \bm{proto}_{i,c})}}
\end{equation}

In Eq.~\eqref{eq7}, we can observe that: 1) by using $M_{g}$, the inter-image diversity is measured by the similarity between the prototypes instead of instance-level pair-wise comparison, thereby remarkably reducing the computational complexity, and 2) $M_{g}(I_{i}, [C])$ encodes the similarity between intra-class prototypes across images and increases if $I_{i}$ is more similar to the picked image set $\Delta{\mathcal{S}}$. 

Based on the observations above, we thus adopt the following \emph{intra-class redundancy rejection} process to enhance the intra-class diversity across images: reject the image $I_{i}$ when $M_{g}(I_{i}, [C])$ is larger than a threshold $T_{intra}$ (0.7 by default), and accept otherwise.

\textbf{Inter-class Diversity.} Though the intra-class diversity can be enhanced based on $M_{g}(I_{i}, [C])$, the image set acquired by the intra-class rejection process tends to favor certain classes (\emph{i.e.} the majority classes), leading to severe class imbalance. To deal with this issue, we increase the inter-class diversity by introducing the \emph{inter-class balancing} process, \emph{i.e.} adaptively providing more budgets for the minority classes than the majority ones. 

Concretely, we first build the minority class set $[C_{minor}]$ by sorting the overall classes according to the frequency of occurrences in the labeled image set $\mathcal{S}$ and selecting the classes with the $C_{minor}$ fewest instances, where $C_{minor}=\alpha C$ ($0<\alpha<1$). We assign each minority class $c\in [C_{minor}] $ a relatively large quota $q_{c}=\frac{\beta}{\alpha{C}}b$ ($\alpha<\beta<1$) as the class-specific budget. 

For an unlabeled image $I_{i}$, we check if it contains the instances from the minority classes by computing
\begin{equation}
\label{eq9}
{M}_{p}(I_{i}, [C_{minor}])=\max\limits_{c\in[C_{minor}]}p(i,c),
\end{equation}
where $p(i,c)=\max\limits_{k\in[t]}{\mathbbm{1}(c,c_{i,k})\cdot{p_{i,k}}}$ estimates the probability about the existence of instances from the class $c$ in $I_{i}$. 

\begin{figure*}[!thp]
	\centering
	\subfloat[FRCNN on MS COCO]{
		\includegraphics[width=0.32\textwidth]{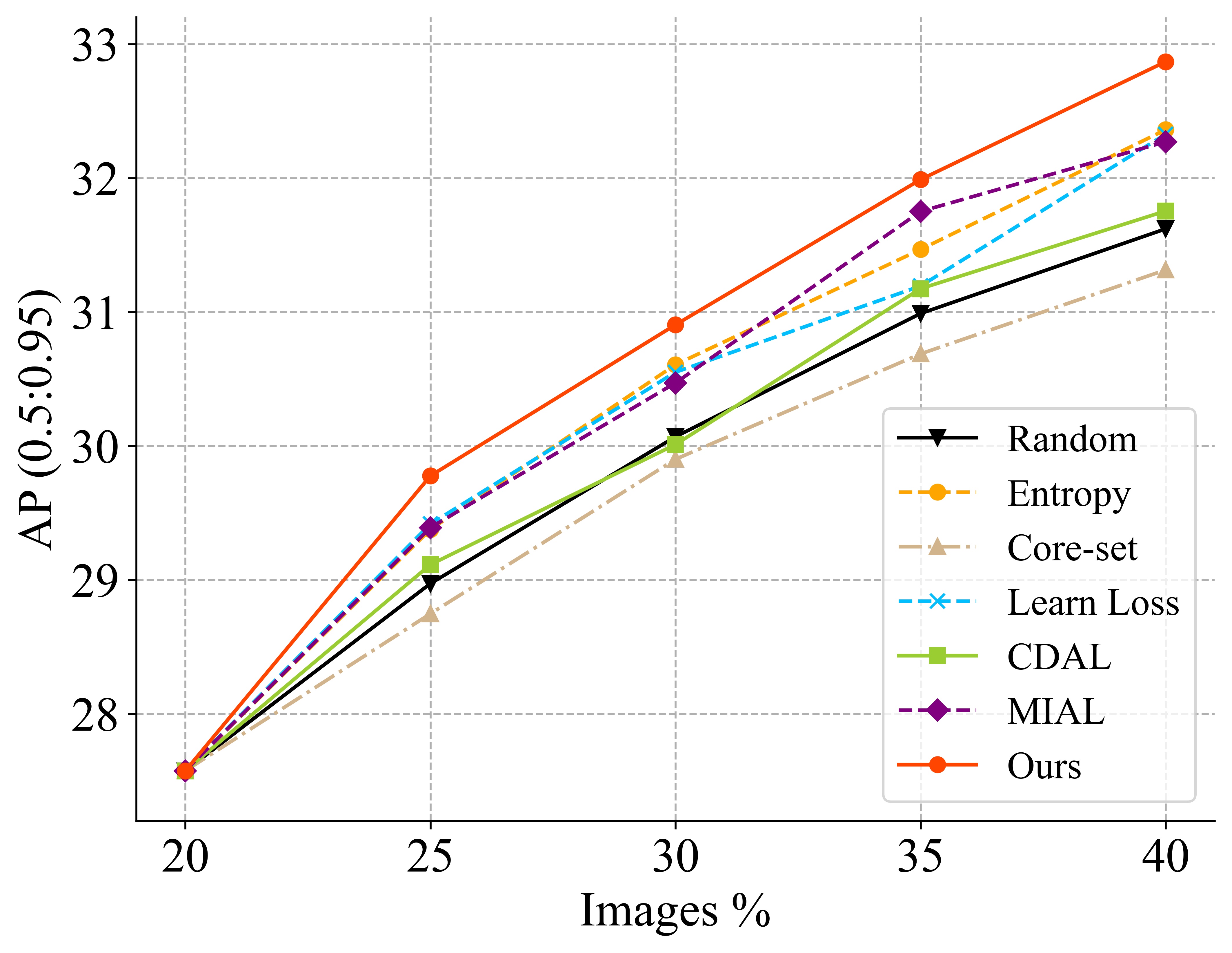}
	}
	\subfloat[RetinaNet on MS COCO]{
		\includegraphics[width=0.32\textwidth]{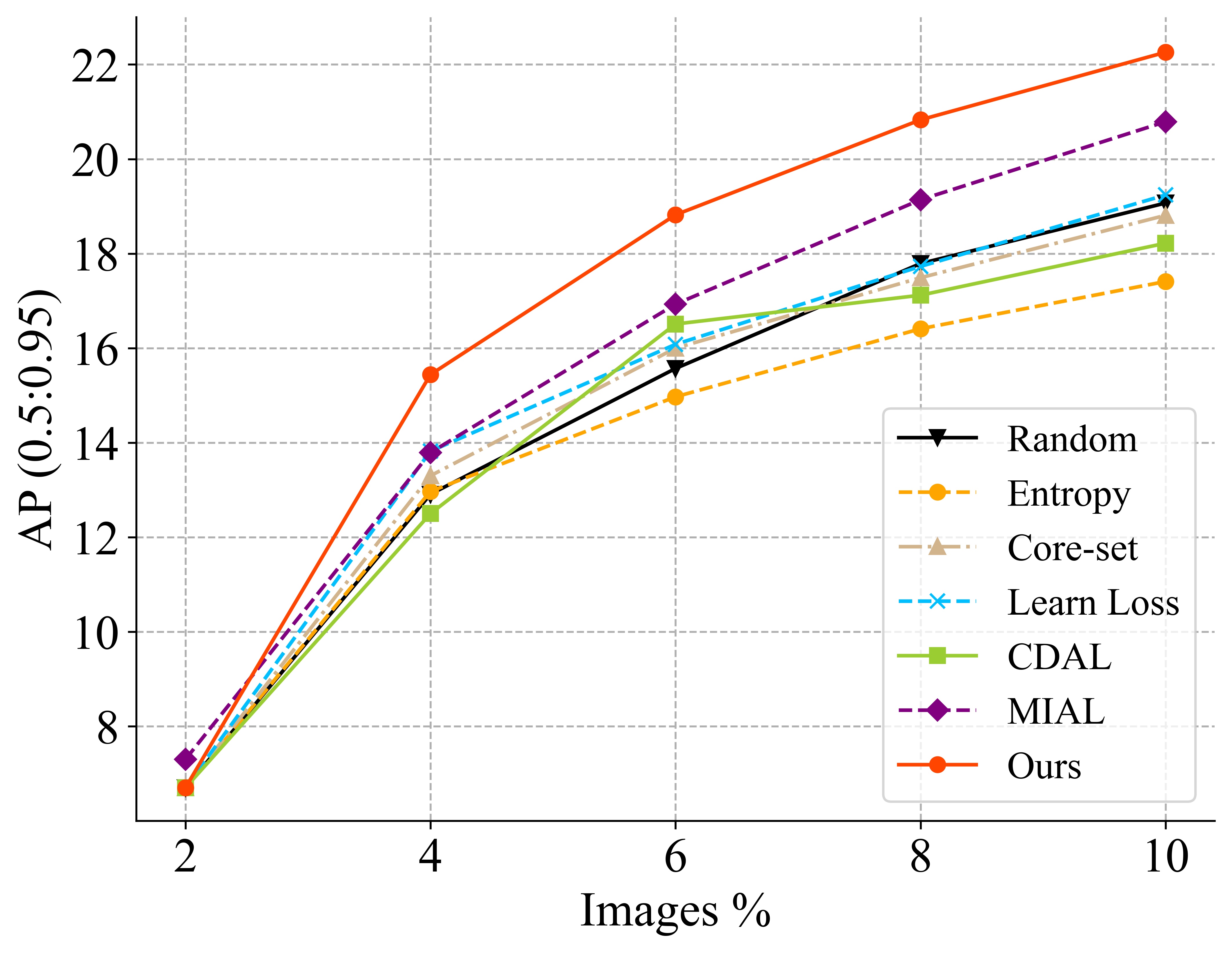}
	}
	\subfloat[SSD on Pascal VOC]{
		\includegraphics[width=0.32\textwidth]{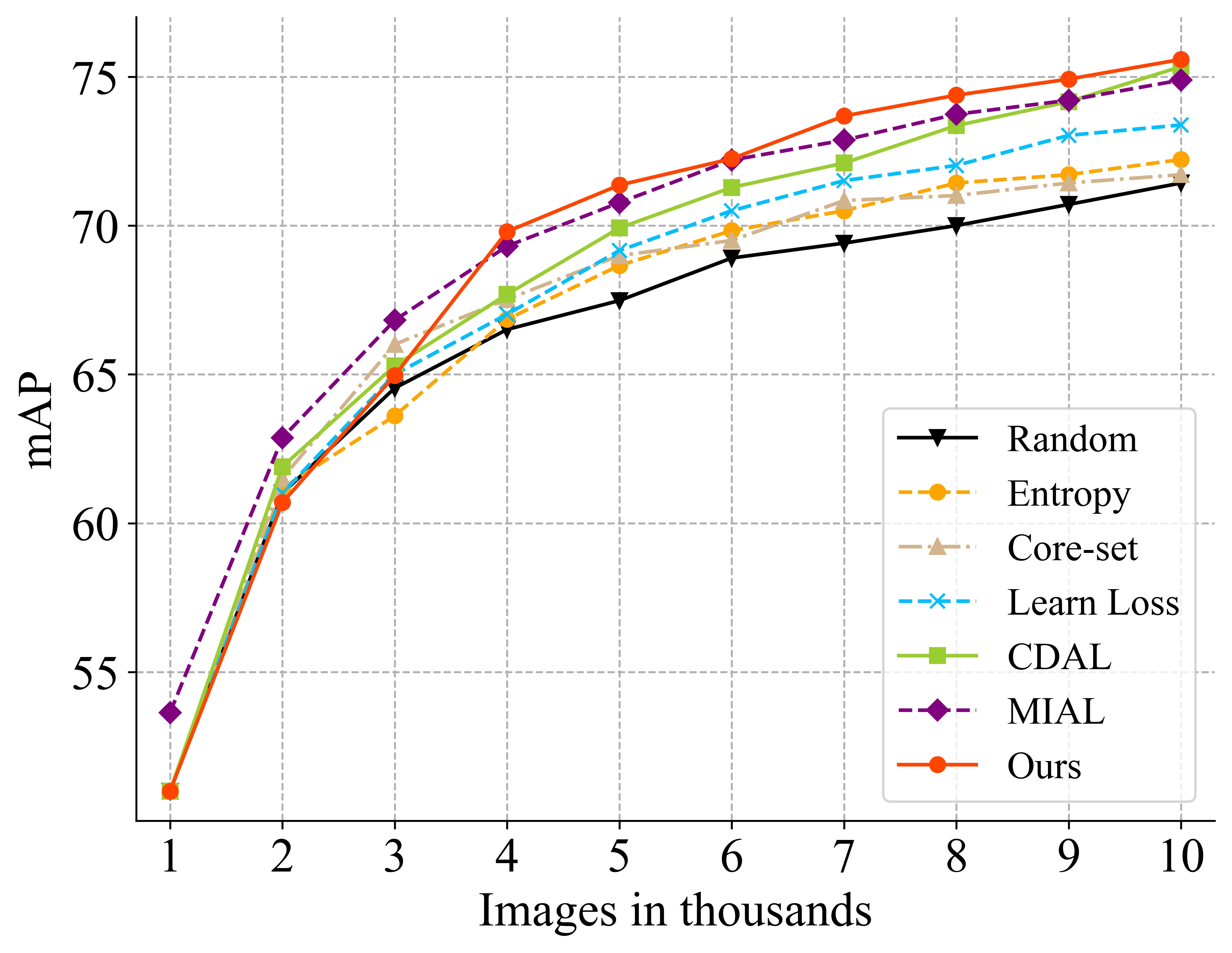}
	}
	\caption{Comparison results. (a)/(b) AP (\%) on MS COCO by using different portions of training data; (c) mAP (\%) on Pascal VOC. (a), (b) and (c) adopt Faster R-CNN and RetinaNet with ResNet-50, SSD with VGG-16, respectively.}\label{fig:coco_voc_results} 
\end{figure*}

In this work, we adopt a threshold $T_{inter}$ (0.3 by default), where the image $I_{i}$ is accepted as containing minority classes if ${M}_{p}(I_{i}, [C_{minor}])>T_{inter}$, and is rejected otherwise. Once $I_{i}$ is accepted, the quota $\{q_{c}\}$ will be updated as $q_{c}:=q_{c}-1$ if $p(i,c)>T_{inter}$. During image acquisition, the class with $q_{c}=0$ will be removed from the minority class set $[C_{minor}]$, while the number of the minority classes $C_{minor}$ is updated by $C_{minor}:=C_{minor}-1$. The whole process is terminated when $C_{minor}$ reaches 0. 

Since an image may contain instances from multiple minority classes and $\beta<1$, the number of acquired images by using the inter-class balancing process should not exceed the budget $b$. We therefore fill up $\Delta{\mathcal{S}}$ with the remaining unlabeled images until the budget $b$ runs out.

By performing the process above, we can make a balance w.r.t. the number of instances from various classes, and finally increase the inter-class diversity. The details of DivProto are summarized in Algorithm \ref{code:whole}.


\section{Experiments}

\subsection{Experimental Settings}
\noindent\textbf{Datasets.} 
We evaluate the proposed method on two benchmarks for object detection: MS COCO~\cite{coco} and Pascal VOC~\cite{voc07,voc12}.
\textbf{MS COCO} has 80 object categories with 118,287 images for training and 5,000 images for validation~\cite{coco}. Similar to \cite{vaal} in dealing with large-scale data, we report the performance with 20\%, 25\%, 30\%, 35\%, 40\% of the training set, where the first 20\% subset is randomly collected. At each acquisition cycle, after the detector is fully trained, 5\% (\emph{i.e.} 5,914) of the total images are acquired from the rest unlabeled set via active learning for annotation. We adopt the Average Precision (AP) over IoU thresholds ranging from 0.5 to 0.95 as the evaluation metric. \textbf{Pascal VOC} contains 20 object categories, consisting of the VOC 2007 trainval set, the VOC 2012 trainval set and the VOC 2007 test set. By following the settings~\cite{ll4al}, we  combine the trainval sets with 16,511 images as unlabeled images, and randomly select 1,000 images as the initial labeled subset. The budget at each acquisition cycle is fixed to 1,000. The mean Average Precision (mAP) at the 0.5 IoU threshold is used as the evaluation metric.

\noindent\textbf{Implementation Details.}
We set $T_{enms}$ for instance-level redundancy removal in ENMS, $T_{intra}$ for the intra-class diversity and $T_{inter}$ for the inter-class diversity to 0.5, 0.7 and 0.3, respectively. $\alpha$ and $\beta$ are set to 0.5 and 0.75, ensuring that at least 75\% budgets are assigned to 50\% of the classes (minority classes).
We utilize Faster R-CNN~\cite{fasterr-cnn} and RetinaNet~\cite{retinanet} with ResNet-50~\cite{resnet} and FPN~\cite{fpn} as the detection models on MS COCO. At all active learning cycles, we train the detector for 12 epochs with batch size 16. The learning rate is initialized as 0.02 and is reduced to 0.002 and 0.0002 after $2/3$ and $8/9$ of the maximal training epoch. On Pascal VOC, we adopt the settings in~\cite{ll4al} using SSD~\cite{ssd} with VGG-16~\cite{vgg} as the base detector. 

\noindent\textbf{Counterpart Methods.}
We make comparison to the state-of-the-art methods. The diversity-based ones include Core-set~\cite{coreset} and CDAL~\cite{cdal}.
To adapt Core-set~\cite{coreset} to the detection task, we follow Learn Loss~\cite{ll4al} to perform $k$-Center-Greedy over the image-level features. 
In regard of CDAL~\cite{cdal}, we apply the reinforcement learning policy on the features after the softmax layer.
The uncertainty-based methods contain Learn Loss~\cite{ll4al} and MIAL~\cite{mial}.
We follow Learn Loss~\cite{ll4al} to add a loss prediction module to simultaneously predict the classification and regression losses. The loss prediction module is trained by comparing image pairs, which empirically performs better than the mean square error~\cite{ll4al}. 
Since loss prediction can affect detector training, we separately conduct active acquisition and detector retraining for fair comparison.

\begin{table*}[!thp]\small
	\centering
	\begin{center}
		\begin{tabular}{c|ccc|ccccc}
			\hline
			\multirow{2}{*}{Method} & \multirow{2}{*}{Entropy} & \multirow{2}{*}{ENMS} & \multirow{2}{*}{DivProto} & \multicolumn{5}{c}{Annotated Percentage} \\ \cline{5-9} 
			&  &  &  & 20\% & 25\% & 30\% & 35\% & 40\% \\ \hline
			Random &  &  &  & 27.57$\pm$0.18 & 28.97$\pm$0.12 & 30.07$\pm$0.24 & 30.99$\pm$0.12 & 31.62$\pm$0.29 \\ \hline
			\multirow{4}{*}{Ours} & $\checkmark$ &  &  & 27.57$\pm$0.18 & 29.38$\pm$0.13 & 30.61$\pm$0.12 & 31.47$\pm$0.17 & 32.36$\pm$0.07 \\
			& $\checkmark$ & $\checkmark$ &  & 27.57$\pm$0.18 & 29.76$\pm$0.16 & 30.82$\pm$0.23 & 31.79$\pm$0.15 & 32.56$\pm$0.09 \\
			& $\checkmark$ &  & $\checkmark$ & 27.57$\pm$0.18 & 29.73$\pm$0.16 & 30.64$\pm$0.11 & 31.86$\pm$0.09 & 32.53$\pm$0.14 \\
			& $\checkmark$ & $\checkmark$ & $\checkmark$ & 27.57$\pm$0.18 & \textbf{29.78}$\pm$0.06 & \textbf{30.90}$\pm$0.14 & \textbf{31.99}$\pm$0.05 & \textbf{32.87}$\pm$0.04 \\ \hline
		\end{tabular}
		\caption{AP (\%) by using different components of the proposed method with Faster R-CNN (ResNet-50) on MS COCO. ``Random" refers to uniform acquisition. With various active acquisition strategies applied, the results are reported with standard deviation over 5 trials.}
		\label{table:ablationmodule}
	\end{center}
\end{table*}

\subsection{Experimental Results}\label{sec:results}


\noindent\textbf{On MS COCO.}
The comparison results on MS COCO are summarized in Fig.~\ref{fig:coco_voc_results}~(a).
The detector built with the complete (100\%) training set achieves 36.8\% AP according to the open-source implementation$\footnote{https://github.com/facebookresearch/maskrcnn-benchmark}$, which can be treated as an approximated upper bound. 
As demonstrated, our method consistently reaches the best performance in all active learning cycles, showing the superiority of the proposed acquisition strategy. In the last cycle with 40\% annotated images, our method achieves 32.87\% AP with an increase of 1.25\%, compared to the uniform random sampling. The uncertainty based methods, \emph{i.e.} Learn Loss~\cite{ll4al} and MIAL~\cite{mial}, deliver almost the same performance as the basic entropy. The diversity-based ones, \emph{i.e.} Core-set~\cite{coco} and CDAL~\cite{cdal}, perform poorly, since they utilize holistic features after spatial pooling without aggregating instance-level information.

\begin{table}[!t]\small
	\centering
	\begin{tabular}{cc|ccc}
		\hline
		\multicolumn{1}{c}{$\alpha$} & $\beta$ & AP & AP$_{50}$ & AP$_{75}$ \\ \hline
		0.50 & 0.25 & 30.68 & 52.48 & 31.93 \\
		0.50 & 0.50 & 30.74 & 52.97 & 31.86 \\
		0.50 & 0.75 & \textbf{30.90} & 53.08 & 32.01 \\
		0.50 & 1.00 & 30.79 & 53.00 & 32.01 \\
		0.25 & 0.75 & 30.71 & 52.90 & 31.63 \\
		0.75 & 0.75 & 30.58 & 52.62 & 32.15 \\ \hline
	\end{tabular}
	\caption{Results at the 30\% cycle using Faster R-CNN (with the ResNet-50 backbone) on MS COCO with various $\alpha$ and $\beta$. AP$_{50}$ (\%) and AP$_{75}$ (\%) refer to AP at the IoU thresholds 0.5 and 0.75, respectively.}
	\label{table:ablahyper}
\end{table}
\begin{table}[!t]\small
	\centering
	\begin{tabular}{c|ccccc}
		\hline
		\multirow{2}{*}{Method} & \multicolumn{5}{c}{Annotated Percentage} \\ \cline{2-6} 
		& 20\% & 25\% & 30\% & 35\% & 40\% \\ \hline
		Random & 29.38 & 31.03 & 32.10 & 33.13 & 33.58 \\
		Entropy & 29.38 & 31.48 & 32.53 & 33.98 & 34.12 \\
		Ours & 29.38 & \textbf{31.79} & \textbf{32.95} & \textbf{34.14} & \textbf{34.89} \\ \hline
	\end{tabular}
	\caption{AP (\%) using Faster R-CNN (ResNet-101) on MS COCO.}
	\label{table:res101}
\end{table}

Additionally, we report the results with small budgets (no more than 10\%) following MIAL~\cite{mial}. As in Fig.~\ref{fig:coco_voc_results}~(b), our method still outperforms the counterpart methods by a large margin in those settings. 
Though the detection performance under such small budgets does not meet the level of real-world applications, the remarkable gains compared with MIAL~\cite{mial} demonstrate its effectiveness. 

\noindent\textbf{On Pascal VOC.}
We follow the same settings and open-source implementation$\footnote{https://github.com/amdegroot/ssd.pytorch}$ as reported in previous studies\cite{ll4al,cdal,mial}, whose result is 77.43\% mAP with all (100\%) training images.
As shown in Fig.~\ref{fig:coco_voc_results}~(b), our method achieves better results than the other counterparts among the 4$k$ to 10$k$ cycles. Note that our method reaches a 73.68\% mAP by using only 7k images. As a contrast, CDAL~\cite{cdal} and MIAL~\cite{mial} need 8k and Learn Loss~\cite{ll4al} needs 10k, showing the advantage of our method in regard of saving annotations. 
Indeed, our method does not perform better at the 2$k$ and 3$k$ cycles. It happens since the detectors are limited due to insufficient training at the initial cycles and cannot distinguish the difference between uncertain queries. This motivates us that the uncertainty and diversity should have varying weights during different active learning periods, but we do not go further here in case of increasing the complexity.
MIAL~\cite{mial} achieves the best performance when using 1k, 2k and 3k images. It should be noted that MIAL performs semi-supervised learning with unlabeled images, which is unfair for comparison, especially when labeled images are far fewer than the unlabeled.
On the contrary, our method focuses only on active learning and the semi-supervised part is not introduced temporarily.

\subsection{Ablation Study}

\noindent\textbf{On ENMS and DivProto.} As shown in Table~\ref{table:ablationmodule},
the baseline entropy-based methods (Faster R-CNN with ResNet-50) separately applying ENMS or DivProto outperform that using uniform sampling and entropy only, showing the superiority by ensuring the diversity at the instance level. A combination of both the modules further boosts the overall accuracy, indicating that the intra-image diversity and inter-image diversity provide complementary diversity constraints for performance improvement.


\noindent\textbf{On hyper-parameters in DivProto.}
$\alpha$ and $\beta$ control the instance-level balance of classes, which is important to the inter-class diversity. We study the effect of these two hyper-parameters at the 30\% cycle on MS COCO, where the same Faster R-CNN detector is used. As summarized in Table~\ref{table:ablahyper}, our method achieves the best performance when $\alpha=0.5$ and $\beta=0.75$.
\noindent\textbf{On different backbones.}
To evaluate the effect of backbones on the detection accuracy, 
we report the performance on Faster R-CNN by using ResNet-101. As shown in Table~\ref{table:res101}, ResNet-101 can generally improve the performance, since it is a stronger backbone compared to ResNet-50. Our active acquisition strategy still consistently outperforms the random uniform sampling and basic entropy, showing its effectiveness with various backbones.

\subsection{Analysis}

\noindent\textbf{Computational complexity.}
We introduce the diversity into the uncertainty-based solution, and the computational complexity and time cost grow accordingly. Thanks to the design ensuring the diversity at the instance level, we reduce the massive computations by converting them into all the three steps of active learning for object detection and thus avoid remarkable complexity increase. The time cost of the diversity-based methods are reported in Table~\ref{table:time}, evaluated on a server with 8 NVIDIA 1080Ti GPUs. As shown in Table~\ref{table:time}, the time cost is unacceptable when comparing each predicted instance pairs (UB), where exponential time increase occurs as each image usually contains multiple objects. By contrast, our proposed method implements the instance-level diversity constraints through the progressive framework and significantly reduces the time cost. Besides, our method spends less time than CDAL~\cite{cdal}, since it requires REINFORCE based model training.


\begin{table}[!t]\small
\centering
\begin{tabular}{c|c|c|c}
\hline
Method & Inference (s) & Acquisition (s) & Full (s)\\ \hline
CDAL~\cite{cdal} & 5,599 & 2,798 & 8,397\\
UB & 1,666 & $9.95\times{10{^7}}$ & $9.96\times{10{^7}}$\\
Ours & 1,702 & 1,015 & 2,717\\ \hline
\end{tabular}
\caption{Comparison of time cost for active learning on MS COCO at the 20\% cycle. ``Inference" refers to prediction on unlabeled images and ``Acquisition" refers to image selection. UB denotes the upper bound of raw instance-level diversity computation.}\label{table:time}
\end{table}

\begin{figure}[!t]
	\centering
	\includegraphics[width=0.47\textwidth]{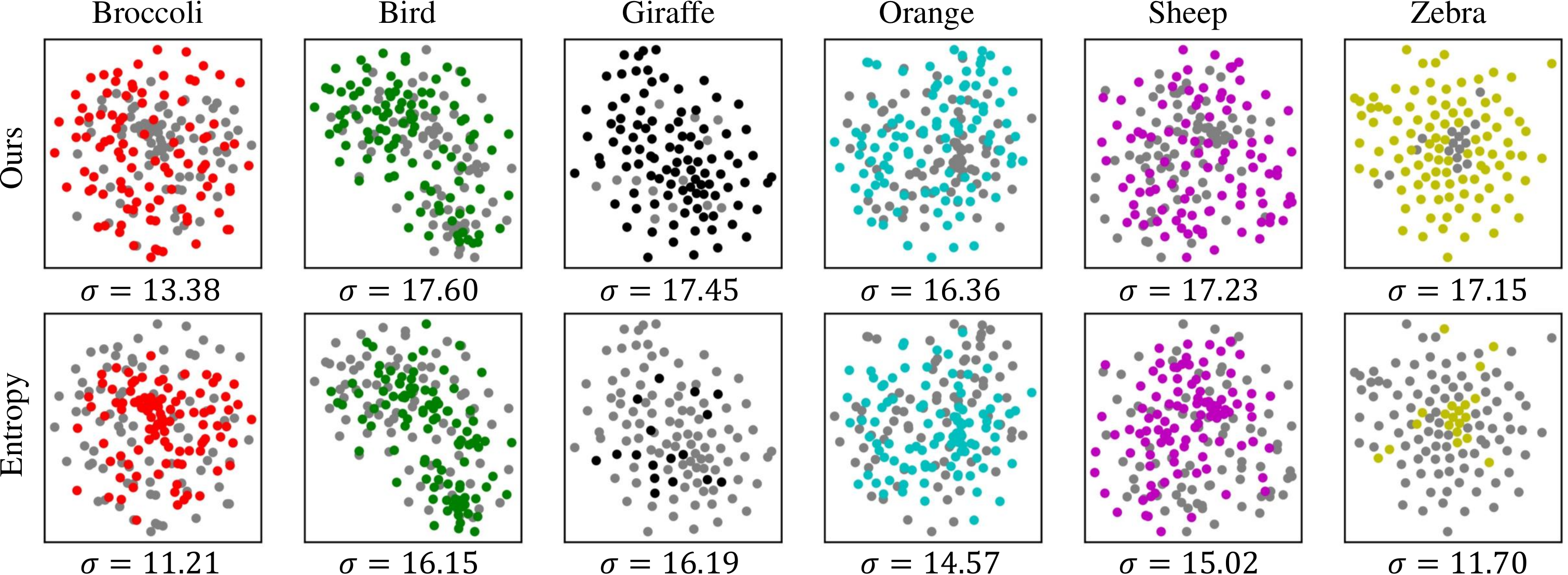}
	\caption{Visualization of the prototypes on MS COCO for 6 classes via t-SNE. Gray points indicate the unselected prototypes. The top and bottom rows are the results by using our method and the Entropy baseline.}
	\label{fig:proto}
\end{figure}
\noindent\textbf{Visualization of prototypes.}
We qualitatively evaluate our method on enhancing the diversity in the presence of prototypes. We choose the basic entropy as the baseline and visualize the prototypes of six categories from MS COCO via t-SNE~\cite{tsne}. We also report the standard deviation $\sigma$ accordingly. 
As illustrated in Fig.~\ref{fig:proto}, the prototypes obtained by DivProto are more representative to the whole unlabeled dataset.
Besides, these prototypes are more diverse according to the standard deviation.

\begin{figure}[!t]
    \centering
    \includegraphics[width=1\linewidth]{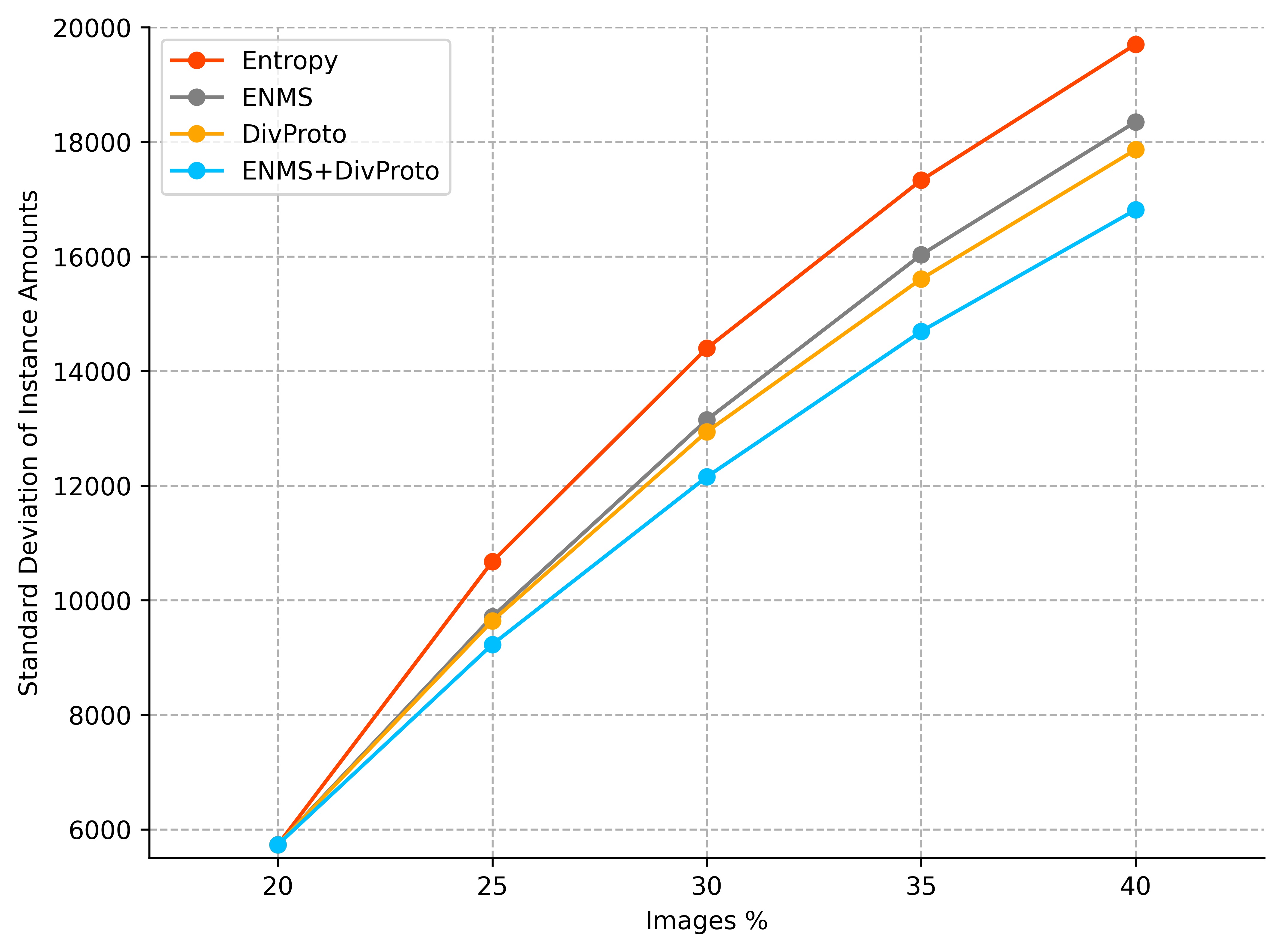}
    \caption{Curves in terms of the standard deviation of instance amounts for 80 classes on MS COCO. Statistics are made on the active acquisition subsets.}
    \label{fig:cls_std}
    \vspace{-0.4cm}
\end{figure}

\noindent\textbf{Discussion on inter-class diversity.}
The active acquisition subsets of MS COCO can be used to make more evaluation on our method. For instance, as shown in Fig.~\ref{fig:cls_std}, we calculate the standard deviations of instance amounts for 80 classes from the subsets, to analyze the inter-class diversity. As illustrated, both the proposed ENMS and DivProto modules decrease the standard deviation, indicating that they perform better in selecting class-balanced subsets of images compared to the basic entropy. A combination of ENMS and DivProto further improves the overall performance, confirming that our method improves the inter-class diversity and helps to construct a balanced subset for stronger detectors.

\section{Conclusion}
In this paper, we propose a novel hybrid active learning method for object detection, which combines the instance-level uncertainty and diversity in a bottom-up manner. ENMS is presented to estimate the instance-level uncertainty for a single image, while DivProto is developed to enhance both the intra-class and inter-class diversities by employing the entropy-based class-specific prototypes. Experimental results achieved on MS COCO and Pascal VOC show that our method outperforms the state of the arts. 

\noindent\textbf{Acknowledgment}
This work is partly supported by the National Natural Science Foundation of China (No. 62022011), the Research Program of State Key Laboratory of Software Development Environment (SKLSDE-2021ZX-04), and the Fundamental Research Funds for the Central Universities.

{\small
\bibliographystyle{ieee_fullname}
\bibliography{main}
}

\clearpage
\appendix
\setcounter{section}{0}
\setcounter{table}{0}
\setcounter{figure}{0}
\renewcommand{\thesection}{\Alph{section}}
\renewcommand{\thefigure}{\Alph{figure}}
\renewcommand{\thetable}{\Alph{table}}

\noindent{\large\textbf{Appendix}}

In this supplementary material, we elaborately analyze how different components affect the performance of the proposed diverse prototype (DivProto) strategy. In addition, we provide comparison results with the latest MDN~\cite{b} and some exploration results with a state-of-the-art detector and semi-supervised learning. To make a more comprehensive evaluation, we present additional experimental results on MS COCO as complements to Fig.~3 of the main paper.

\section{Ablation Study of DivProto}
As depicted in Section 4.3, DivProto consists of intra-class redundancy rejection and inter-class balanced selection. We separately evaluate the effects of these two parts, with the Basic Entropy as the baseline for comparison. As summarized in Table~\ref{table:abla_divproto}, both intra-class rejection and inter-class balanced selection improve the performance of the baseline under various annotation percentages. Their combination further promotes the AP with 25\% and 35\% annotated percentages and remains highly competitive in the other cases.

\begin{table}[!h]\small
	\centering
	\begin{tabular}{c|ccccc}
		\hline
		\multirow{2}{*}{Method} & \multicolumn{5}{c}{Annotated Percentage} \\ \cline{2-6} 
		& 20\% & 25\% & 30\% & 35\% & 40\% \\ \hline
		Basic Entropy & 27.57 & 29.38 & 30.61 & 31.47 & 32.36 \\
		Intra-class & 27.57 & 29.52 & 30.32 & 31.15 & \textbf{32.57} \\
		Inter-class & 27.57 & 29.59 & \textbf{30.70} & 31.83 & 32.37 \\
		Both & 27.57 & \textbf{29.73} & 30.64 & \textbf{31.86} & 32.53 \\ \hline
	\end{tabular}
	\caption{AP (\%) on MS COCO by using intra-class redundancy rejection, inter-class balanced selection and their combination, compared to the Basic Entropy baseline. All the methods are based on Faster R-CNN with the ResNet-50 backbone. The best result for each method is highlighted in bold.}
	\label{table:abla_divproto}
\end{table}

\section{Comparison with the latest work MDN}\label{sec:iccv}

MDN~\cite{b} delivers gains in two ways: an acquisition method based on uncertainty disentanglement and an improved SSD detector based on GMM. In contrast, our method mainly focuses on acquisition, and ENMS and DivProto are proposed to handle redundant uncertainty estimation and insufficient cross-image diversity, both of which are not considered in~\cite{b}. To eliminate the effect of the detector, we apply our acquisition method to the improved SSD on VOC07+12 with the same setting as \cite{b}. As in Table \ref{table:iccv}, our method outperforms \cite{b}, showing its advantage in acquisition.

\begin{table}[!h]\small
\centering
\footnotesize
\begin{tabular}{c|c|ccc}
\hline
Acquisition Method & Detector & 2k & 3k & 4k \\ \hline
MDN~\cite{b}                & Improved SSD~\cite{b}      & 61.30   & 66.57   & 68.49   \\
\textbf{Ours}               & Improved SSD~\cite{b}      & 63.35   & 67.56   & 70.33   \\ \hline
\end{tabular}
\caption{mAP (\%) of different acquisition methods on VOC07+12.}
\label{table:iccv}
\end{table}

\section{Results on SOTA Detectors}
Our method is designed for active acquisition independent of detectors, thus theoretically being effective for different detectors. We additionally evaluate our method by using YOLOv5$\footnote{https://github.com/ultralytics/yolov5}$ in Table \ref{table:yolo}, showing its effectiveness with SOTA detectors.

\begin{table}[!h]\small
\centering
\footnotesize
\begin{tabular}{c|ccccc}
\hline
Method & 2\% & 4\%   & 6\%   & 8\%   & 10\%  \\ \hline
Random & 9.96                     & 15.93 & 19.96 & 23.37 & 25.06 \\
\textbf{Ours}   & 9.96                     & 16.68 & 21.03 & 24.06 & 26.39 \\ \hline
\end{tabular}
\caption{AP (\%) using the YOLOv5 detector on MS COCO.}
\label{table:yolo}
\end{table}

\section{Combining with Semi-supervised Learning}

In our opinion, semi-supervised learning is similar to active learning in the goal of pursuing better performance with less annotated data but in different learning paradigms. As in Table \ref{table:ubt}, we achieve better results by combining our method with a reputed semi-supervised one, UBT~\cite{ubt}, where they complement each other.

\begin{table}[!h]\small
\centering
\footnotesize
\begin{tabular}{c|ccccc}
\hline
Method & 2\% & 4\%   & 6\%   & 8\%   & 10\%  \\ \hline
\textbf{Ours}   & 6.70 & 15.44 & 18.82 & 20.83 & 22.26 \\
UBT & 24.30 & 27.01 & 28.45 & 29.41 & 31.50 \\
\textbf{Ours}+UBT & 24.30 & 28.55 & 30.28 & 31.70 & 32.24 \\
\hline
\end{tabular}
\caption{AP (\%) by combining UBT on MS COCO.}
\label{table:ubt}
\end{table}

\section{Comprehensive Results on MS COCO}

In Fig.~3 of the main paper, we report the Average Precision (AP) over IoU thresholds from 0.5 to 0.95 on MS COCO, by using our method as well as the counterpart methods: Core-set~\cite{coreset}, CDAL~\cite{cdal}, Learn Loss~\cite{ll4al}, and MIAL~\cite{mial}. In Table~\ref{table:cocoresult}, we provide more comparison results under different metrics, including AP$_{50}$, AP$_{75}$, AP$_{S}$, AP$_{M}$, and AP$_{L}$. Here, AP$_{50}$, AP$_{75}$ are AP at the 0.5 and 0.75 IoU thresholds, respectively. AP$_{S}$, AP$_{M}$, and AP$_{L}$ indicate AP for small, medium, and large objects, respectively.

As shown in Table~\ref{table:cocoresult}, our method remarkably outperforms the counterparts in most cases. It is worth noting that Core-set~\cite{coreset} performs better than ours at detecting large objects, since it adopts spatial pooling to merge instance-level features to a holistic image-level representation, based on which the significance of large objects is strengthened. By contrast, our method employs a more balanced way to integrate instance-level features, thus achieving a higher averaged AP for all scales at the cost of slightly lower AP for large objects. 



\begin{table*}[!t]
	\centering
	\begin{tabular}{c|c|ccc|ccc}
		\hline
		\multicolumn{1}{c|}{Images} & Method & AP & AP$_{50}$ & AP$_{75}$ & AP$_{S}$ & AP$_{M}$ & AP$_{L}$ \\ \hline
		20\% & Random & 27.57 & 48.81 & 28.09 & 14.53 & 29.87 & 36.12 \\ \hline
		\multirow{9}{*}{25\%} & Random & 28.97 & 50.43 & 29.84 & 15.39 & 31.25 & 37.84 \\
		& Core-set & 28.75 & 50.00 & 29.56 & 14.59 & 30.98 & \textbf{38.79} \\
		& CDAL & 29.11 & 48.60 & 27.86 & 14.29 & 29.69 & 35.82 \\
		& Learn Loss & 29.42 & 51.08 & 30.38 & 16.35 & 31.96 & 37.81 \\
		& MIAL & 29.39 & 51.21 & 30.36 & 15.96 & 31.87 & 38.68 \\
		& Entropy & 29.38 & 51.06 & 30.31 & 16.18 & 31.63 & 37.89 \\
		& ENMS & 29.76 & 51.64 & 30.78 & \textbf{16.72} & 32.12 & 38.03 \\
		& DivProto & 29.73 & 51.45 & \textbf{30.84} & 16.62 & 32.32 & 38.08 \\
		& Ours & \textbf{29.78} & \textbf{51.70} & 30.81 & 16.52 & \textbf{32.32} & 38.00 \\ \hline
		\multirow{9}{*}{30\%} & Random & 30.07 & 51.61 & 31.13 & 16.05 & 32.42 & 39.42 \\
		& Core-set & 29.90 & 51.37 & 31.12 & 15.60 & 32.19 & \textbf{40.65} \\
		& CDAL & 30.01 & 51.57 & 31.27 & 16.28 & 32.72 & 39.21 \\
		& Learn Loss & 30.55 & 52.53 & 31.90 & \textbf{17.68} & 33.10 & 38.52 \\
		& MIAL & 30.47 & 52.49 & 31.45 & 16.85 & 33.44 & 39.31 \\
		& Entropy & 30.61 & 52.66 & 31.96 & 17.49 & 33.13 & 38.97 \\
		& ENMS & 30.82 & 52.89 & \textbf{32.07} & 17.40 & \textbf{33.48} & 38.88 \\
		& DivProto & 30.64 & 52.64 & 31.77 & 17.21 & 33.37 & 38.92 \\
		& Ours & \textbf{30.90} & \textbf{53.08} & 32.01 & 17.56 & 33.44 & 39.10 \\ \hline
		\multirow{8}{*}{35\%} & Random & 30.99 & 52.70 & 32.47 & 16.88 & 33.52 & 40.35 \\
		& Core-set & 30.69 & 52.25 & 31.97 & 15.96 & 33.17 & \textbf{41.76} \\
		& CDAL & 31.17 & 53.07 & 32.55 & 17.22 & 33.84 & 40.56 \\
		& Learn Loss & 31.19 & 53.19 & 32.67 & 17.48 & 34.04 & 39.23 \\
		& MIAL & 31.75 & 53.89 & 33.42 & 17.51 & \textbf{34.60} & 41.09 \\
		& Entropy & 31.47 & 53.59 & 32.95 & 18.01 & 34.16 & 39.76 \\
		& ENMS & 31.79 & 54.05 & 33.29 & 18.11 & 34.50 & 40.14 \\
		& DivProto & 31.86 & 54.08 & 33.41 & \textbf{18.24} & 34.58 & 40.56 \\
		& Ours & \textbf{31.99} & \textbf{54.18} & \textbf{33.51} & 18.09 & 34.39 & 40.54 \\ \hline
		\multirow{8}{*}{40\%} & Random & 31.62 & 53.29 & 33.18 & 17.14 & 34.19 & 41.26 \\
		& Core-set & 31.31 & 52.89 & 32.80 & 16.19 & 33.81 & \textbf{42.85} \\
		& CDAL & 31.75 & 53.57 & 33.38 & 17.75 & 34.53 & 41.23 \\
		& Learn Loss & 32.33 & 54.61 & 33.92 & 18.72 & 35.37 & 40.76 \\
		& MIAL & 32.27 & 54.69 & 34.04 & 17.72 & 35.27 & 41.86 \\
		& Entropy & 32.36 & 54.69 & 33.97 & 18.67 & 35.25 & 40.63 \\
		& ENMS & 32.56 & 54.84 & 34.25 & 18.53 & 35.42 & 40.96 \\
		& DivProto & 32.53 & 54.77 & 34.24 & 18.57 & 35.34 & 41.44 \\
		& Ours & \textbf{32.87} & \textbf{55.15} & \textbf{34.58} & \textbf{18.99} & \textbf{35.57} & 41.44 \\ \hline
	\end{tabular}
	\caption{AP by using various active learning based methods on MS COCO. All the results are based on Faster R-CNN with the ResNet-50 backbone. 
		AP$_{50}$ and AP$_{75}$ refer to AP at the 0.5 and 0.75 IoU thresholds. AP$_{S}$, AP$_{M}$, and AP$_{L}$ indicate AP for objects with small, medium, and large sizes, respectively.}
	\label{table:cocoresult}
\end{table*}

\end{document}